\documentclass[pdflatex,sn-mathphys]{sn-jnl}

\theoremstyle{thmstyleone}%
\theoremstyle{thmstyletwo}%

\theoremstyle{thmstylethree}%

\usepackage{derivative}
\begin{document}

\title[ ]{TaLU: A Hybrid Activation Function Combining Tanh and Rectified Linear Unit to Enhance Neural Networks}


\author*[1]{\fnm{Md. Mehedi} \sur{Hasan}}\email{mmehedihasann@gmail.com}
\author[1]{\fnm{Md. Ali} \sur{Hossain}}\email{ali.ruet@gmail.com}
\author[1]{\fnm{Azmain Yakin} \sur{Srizon}}\email{azmainsrizon@gmail.com}
\author[1]{\fnm{Abu} \sur{Sayeed}}\email{abusayeed.cse@gmail.com}

\affil[1]{\orgdiv{Department of Computer Science \& Engineering}, \orgname{Rajshahi University of Engineering \& Technology}, \orgaddress{\street{Kazla}, \city{Rajshahi}, \postcode{6204},  \country{Bangladesh}}}


\abstract{The application of the deep learning model in classification plays an important role in the accurate detection of the target objects. However, the accuracy is affected by the activation function in the hidden and output layer. In this paper, an activation function called TaLU, which is a combination of Tanh and Rectified Linear Units (ReLU), is used to improve the 
prediction. ReLU activation function is used by many deep learning researchers for its computational efficiency, ease of implementation, intuitive nature, etc. However, it suffers from a dying gradient problem. For instance, when the input is negative, its output is always zero because its gradient is zero. A number of researchers used different approaches to solve this issue.  Some of the most notable are LeakyReLU, Softplus, Softsign, Elu, ThresholdedReLU, etc. This research developed TaLU, a modified activation function combining Tanh and ReLU, which mitigates the dying gradient problem of ReLU. The deep learning model with the proposed activation function was tested on MNIST and CIFAR-10, and it outperforms ReLU and some other studied activation functions in terms of accuracy(upto 6\% in most cases, when used with Batch Normalization and a reasonable learning rate). }




\maketitle

\section{Introduction}\label{sec1}
In the 1970s, a notable limitation of Artificial Intelligence was observed and it was due to the lack of ability of neural networks to differentiate among multiple classes or even binary classification problems namely XOR problems which couldn't be solved using single neurons \cite{bib1,sri1,sri2}. One of the causes of this problem is the fact that neurons couldn't incorporate non-linearity at that time. Later, this nonlinearity was incorporated into Neurons through the use of activation function \cite{relu,sri3,sri4}. It also controls how the information propagates among different layers of the neural network, especially if it is a deep neural network. Therefore, a better activation function can enable a superior deep neural network. Some deep networks such as Long-Short Term Memory (LSTM) models especially prefer Sigmoid or Tangent activation function \cite{sri5,sri6}. Others like Convolutional Neural Networks (CNN) and Multi-layer perceptrons (MLP) use ReLU \cite{relu} or other activation functions. This research focuses on the ReLU \cite{relu} or similar activation functions to enhance the performance of Convolutional Neural Network (CNN) and Multi-Layer Perceptron (MLP) networks.

The ReLU has proved its worth in various studies and works \cite{relu_usage,relu_usage2,relu_usage3}. As ReLU uses identity in the positive information flow, it allows a much deeper neural network to be trained with ease which makes it a popular choice. Another reason behind its success is the fact that ReLU is computationally efficient since it provides zero negative inputs \cite{sri7,sri8}. It provides a sweet mixture of efficiency and simplicity which researchers cherish \cite{relu}. However, the researchers do not stop finding alternatives to ReLU. The reasons behind this search are two-fold. One is missing negative information and the other is lacking zero-center property \cite{frelu}.

Although, ReLU is computationally efficient and provides sparsity to the neural network, however, it doesn't care about negative values. It simply ignores them and provides zero instead. However, these negative values are essential for learning \cite{frelu}. Therefore, various activation functions like LeakyReLU \cite{leakyrelu}, Parameterized ReLU \cite{prelu}, Randomized ReLU \cite{rrelu}, FReLU \cite{frelu}, ELU \cite{elu}, SELU \cite{selu}, Swish \cite{swish} etc. uses the negative slope into them in some ways.

The authors of ELU \cite{elu} proved that shifting the mean towards zero speeds up learning in the deep neural network. Adding negative values into the activation does shift the mean towards zero means. However, some activation function doesn't work well with Batch Normalization and which is repulsive to the researchers \cite{sri9,sri10}. 

\section{Methodology}
In this section, all related methodology used in this paper are discussed.
\subsection{TaLU}
We can represent ReLU\cite{relu} as:
\begin{equation}\label{eq1}
    ReLU(x)=
    \begin{cases}
      x, & \text{if}\ x>0 \\
      0, & \text{otherwise}
    \end{cases}
  \end{equation}

\begin{figure}[ht]
\centering
\includegraphics[width=1.00\textwidth]{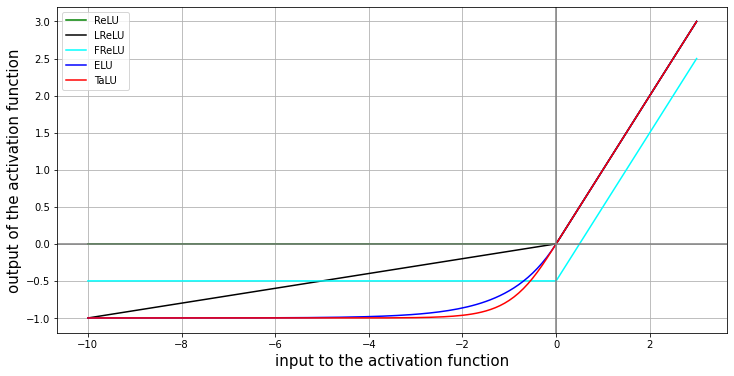}
\caption{\centering TaLU with other activation functions}
\label{fig:fig-1}
\end{figure}

Since ReLU doesn't have any value when encountered with negative input, therefore Tanh activation function is proposed in that scenario, effectively making TaLU, a combination of ReLU and Tanh activation functions and incorporating negative values into the training process. TaLU is represented as:
\begin{equation}\label{eq2}
    TaLU(x)=
    \begin{cases}
      x, & \text{if}\ x>0 \\
      \alpha tanh(x), & \text{otherwise}
    \end{cases}
  \end{equation}
where $\alpha$ is a trainable parameter. The default value of $\alpha$ is 1.0. If we set $\alpha = 0$, we get ReLU with a trainable parameter $\alpha$.

\begin{figure}[ht]
\centering
\includegraphics[width=1.00\textwidth]{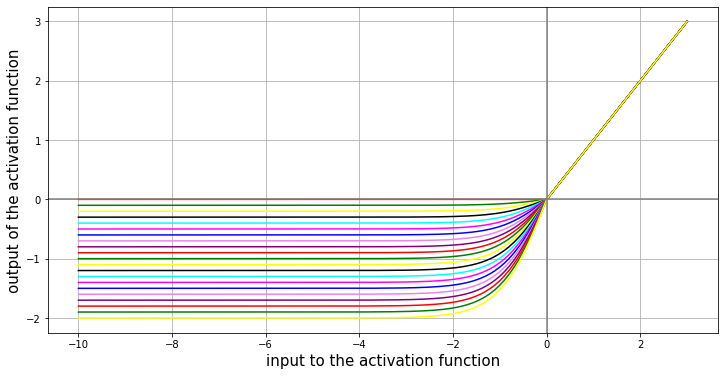}
\caption{\centering From top to bottom: TaLU with 0.0 to 2.0 alpha value, with 0.1 increments}
\label{fig:fig-2}
\end{figure}

This is true for the forward pass of the training process of a neural network. For the backward pass or during backpropagation, TaLU is rewritten as:
\begin{equation}\label{eq3}
    \dfrac{d}{dx}{TaLU(x)}=
    \begin{cases}
      1, & \text{if}\ x>0 \\
       \alpha sech^2({x}), & \text{otherwise}
    \end{cases}
  \end{equation}

\subsubsection{Input to TaLU}
If an image has some pixel having values [-20, -1.0, 0.0, 1.0, 20], ReLU activation function would generate [0.0, 0.0, 0.0, 1.0, 20]. So, a noticeable information is lost in the process. However, using TaLU for the same input pixel values, the output is [-1.0, -0.7615941559557649, 0.0, 1.0, 20]. TaLU retains more information than ReLU and it is better for learning.

\subsection{Compatibility with Batch Normalization}
In the conclusion section of ELU paper \cite{elu} authors have reached the verdict that ELU is not compatible with BatchNormalization \cite{batch}. This is true for some other activation functions like PReLU, and SReLU which is shown by the authors of \cite{frelu}. Therefore, a conclusion is reached because of the conflict between the representation restoration of Batch Normalization and the presence of the negative parameter in the activation function. FReLU is compatible with Batch Normalization as proved by the authors \cite{frelu}. In this paper, we use Batch normalization which stabilizes our training process while using a large learning rate which maximizes performance and converges faster. We show this through our experiment in Table~\ref{res2}.

\subsection{Comparisons among existing activation functions with TaLU}
A comparative study of the ReLU, ELU, LeakyReLU, Swish, SELU, etc. activation functions is provided for better understanding.
In this subsection, we provide a comparison among various correlative activation functions.

\subsubsection{ReLU}
The definition of activation function ReLU which is defined in Equation~\ref{eq1}, maximum of 0 or the input. The proposed activation function is an extension of this equation. Instead of 0 for negative inputs, Tanh function is used with a trainable parameter $\alpha$. This enables TaLU to behave exactly like ReLU if we set $\alpha = 0$. When $\alpha > 0$, TaLU increases the area of expression for the activation function. 

\subsubsection{Leaky ReLU or Parametric ReLU}
This activation function is defined by the equation $PReLU(x) = MAX(0,x) + K * MIN(0,x)$, where K is a trainable parameter. If we fix the K with a sufficiently small value, PReLU effectively becomes LReLU \cite{leakyrelu}.

\subsubsection{SELU}
The SELU \cite{selu} activation function is defined as: $f(x) = scale * (max(0,x) + min(0, alpha * (exp(x) - 1)))$ where x is the input to the activation function, alpha is a scaling parameter, and scale is a normalization factor that helps to keep the mean and variance of the output of each layer constant over the course of training.

The SELU activation function is designed to work well with deep neural networks by maintaining the mean and variance of the output of each layer close to 0 and 1, respectively. This helps to prevent the vanishing gradient problem that can occur in deep neural networks, which can make training difficult.
The ReLU component of the SELU function is used to introduce non-linearity and to activate neurons that have positive inputs, while the Tanh component is used to activate neurons that have negative inputs. The scaling factor alpha is chosen to make the output of the SELU function behave like a Gaussian distribution when the inputs to the function have a zero mean and unit variance.

\subsection{GELU}
The GELU \cite{gelu} activation function can be expressed mathematically as follows:
$GELU(x) = 0.5 * x * (1 + tanh(sqrt(2/pi)*(x+0.044715*x^3)))$
where x is the activation function's input. The GELU function, which helps to solve the vanishing gradient issue, is essentially a smoother variant of the ReLU function. The function can be utilized in gradient-based optimization algorithms because it is created to be continuous and differentiable. The formula makes use of the Gaussian cumulative distribution function to guarantee that, for big negative values, the function remains close to linear and smoothly shifts to a non-linear domain for positive values. Overall, the GELU activation function, which is frequently utilized in deep learning systems, is a promising substitute for ReLU and other activation functions.

\subsubsection{ELU}
ELU \cite{elu} is defined as $ELU(x) = MAX(0,x) + MIN(exp(x) - 1), 0$. Both FReLU and ELU share several characteristics and forms. Both ELU and TaLU use exponential operation, however, FReLU uses a bias term. Experiments show that exponential terms perform well in terms of accuracy despite being expensive in terms of time complexity. Moreover, TaLU and FReLU are more compatible than ELU.

\subsubsection{Swish}
Swish \cite{swish} is defined as $f(x) = x * sigmoid(beta * x)$ where x is the input to the activation function and beta is the hyperparameter that controls the shape of the function. The sigmoid function acts as a gating mechanism, which means that it controls the flow of information through the network by scaling the input x. When x is positive, the sigmoid function scales it up, and when x is negative, the sigmoid function scales it down. This self-gating property allows Swish to adaptively adjust the input to the activation function. One of the main advantages of Swish over other activation functions is its smoothness. Unlike ReLU, which has a sharp corner at zero, Swish is a smooth function that can be easily differentiated. This means that it can be used in deep neural networks without the risk of exploding or vanishing gradients, which can impede training. 

\subsubsection{Softplus}
The ReLU activation function, which is defined as the logarithm of the exponential function plus one, is approximated by the Softplus activation function \cite{softplus}. The following is a written representation of the mathematical expression for the Softplus activation function:
$Softplus(x) = ln(1 + exp(x))$
where x is the activation function's input. Due to its characteristics, the Softplus function—which is continuous and differentiable—is frequently utilized as an activation function in neural networks. The Softplus function is differentiable at x = 0 as opposed to the ReLU function, making it a better option for optimization techniques that demand gradients. Additionally, the Softplus function can be used for binary classification tasks as an alternative to the sigmoid function \cite{sigmoid}. Because the Softplus function is non-monotonic, it can generate both positive and negative results for a particular input. The Softplus activation function has been utilized successfully in a variety of applications, including computer vision and natural language processing, and is an important tool for deep learning practitioners.

\subsection{Softsign}
Similar to the sigmoid function, the Softsign activation function \cite{softsign} is a smooth and bounded activation function. As the ratio of the input to the product of the absolute value of the input plus one, it is defined. The Softsign activation function's mathematical expression is as follows:
$Softsign(x) = x / (1 + |x|)$
where x is the activation function's input. Due to its characteristics, the Softsign function—which is continuous and differentiable—is frequently utilized as an activation function in neural networks. The Softsign function can be helpful for preventing big values from spreading throughout the network during training because its output is constrained. Because the Softsign function is non-monotonic, it can capture input-output relationships that are more complex. The Softsign function, like the sigmoid function, can experience the vanishing gradient problem at extremely high x values. The Softsign activation function has been utilized successfully in a number of applications, including speech recognition and picture classification, and is a useful tool for deep learning practitioners in general.

\begin{figure}[!htbp] 
\centering
\includegraphics[width=1.00\textwidth]{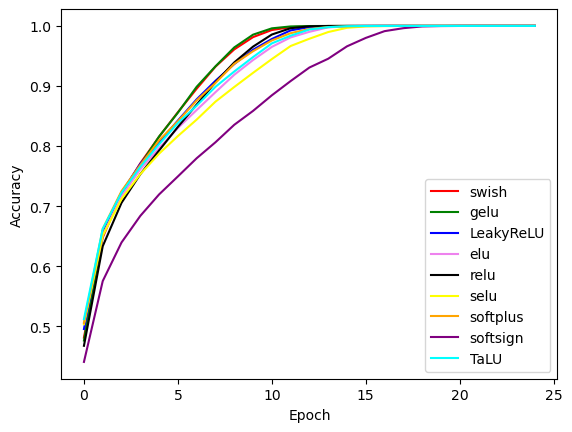}
\caption{\centering Accuracy curve for different activation function for Simple CNN with 0.001 learning rate and 50 epoch usuing CIFAR-10 dataset}
\label{accuracy}
\end{figure}

\begin{figure}[!htbp]
\centering
\includegraphics[width=1.00\textwidth]{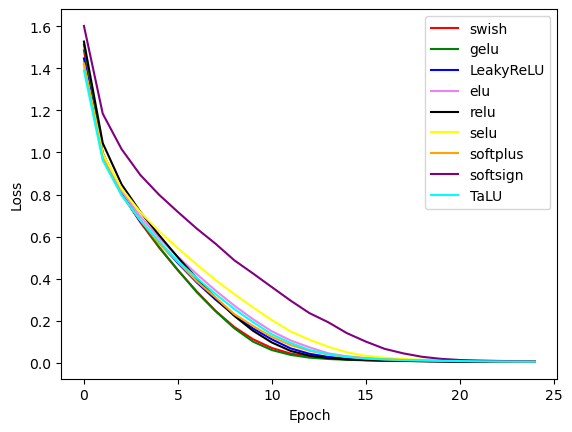}
\caption{\centering Loss curve for different activation function for Simple CNN with 0.001 learning rate and 50 epoch usuing CIFAR-10 dataset}
\label{loss}
\end{figure}

\section{Experimental Setup}
The CIFAR-10 \cite{cifar}, and MNIST \cite{mnist} standard image datasets are used in this experiment for the task of evaluation of TaLU. Using the training settings indicated in this section for the Simple CNN shown in Table~\ref{architecture}, we conduct all of our tests. The starting default learning rate is set to 0.1/0.01. The value of momentum. was set at 0.9.  Models for CIFAR-10 and MNIST are trained using stochastic gradient descent (SGD \cite{sgd}) with 512-batch sizes over 25 iterations (no warming up). Moreover, the initialization value for the TaLU parameter alpha in this paper is 1.0 with trainable alpha parameter by default. All experimental findings across all datasets are shown using default seeds over a single run of 25 epochs.

To compare the proposed activation functions' performance with existing activation function, a set of experiments are designed with 2 types of convolutional neural networks such as a simple one and a residual one respectively. In the experiment, a 8 layer simple convolutional neural network is used which is shown in Table~\ref{architecture}. The same architecture is used for all the comparison for normal convolutional neural network. For residual network, the layer architecture is shown in Table~\ref{architecture2}.

\begin{table}[ht]
\begin{center}
\begin{minipage}{200pt}
\caption{Basic CNN architecture used to compare activation functions. For MNIST, input shape is 28x28x1}\label{architecture}%
\begin{tabular}{@{}lll@{}}
\toprule
Layer & Output Shape  & Parameter \#\\
\midrule
 Conv2D     &      (None, 32, 32, 32)   &     896 \\      
                                                                 
 TaLU        &      (None, 32, 32, 32)   &     1        \\ 
                                                                 
 Conv2D    &      (None, 32, 32, 32)   &     9248      \\
                                                                 
 TaLU       &       (None, 32, 32, 32)   &     1        \\ 
                                                                 
 MaxPooling2D &  (None, 16, 16, 32)   &    0         \\                                                             
                                                                 
 Conv2D   &       (None, 16, 16, 64)    &    18496     \\
                                                                 
 TaLU       &      (None, 16, 16, 64)   &     1         \\
                                                                 
 Conv2D    &       (None, 16, 16, 64)    &    36928   \\  
                                                                 
 TaLU      &       (None, 16, 16, 64)   &     1        \\ 
                                                                 
 MaxPooling2D &  (None, 8, 8, 64)      &   0    \\     
                                                                                                                           
 Conv2D   &       (None, 8, 8, 128)      &   73856    \\ 
                                                                 
 TaLU      &       (None, 8, 8, 128)     &    1      \\   
                                                                 
 Conv2D     &     (None, 8, 8, 128)     &    147584    \\
                                                                 
 TaLU       &      (None, 8, 8, 128)    &     1        \\ 
                                                                 
 MaxPooling2D  & (None, 4, 4, 128)     &   0      \\

 Flatten    &     (None, 2048)         &     0         \\
                                                                 
 Dense      &        (None, 128)      &         262272    \\
                                                                 
 TaLU       &      (None, 128)       &        1        \\ 
                                                                 
 Dense      &       (None, 10)       &         1290      \\
\botrule
\end{tabular}
\footnotetext{
Total params: 550,577\\
Trainable params: 550,577\\
Non-trainable params: 0
}
\end{minipage}
\end{center}
\end{table}

\begin{table}[!htbp]
\begin{center}
\begin{minipage}{\textwidth}
\caption{Basic Residual CNN architecture used to compare activation functions. For MNIST, input shape is 28x28x1}\label{architecture2}%
\begin{tabular}{@{}llll@{}}
\toprule
Layer & Output Shape  & Parameter \# & Connected to\\
\midrule
 InputLayer    &      (None, 32, 32, 3) & 0    &       [] \\                              
                                                                                                  
 Conv2D1      &      (None, 32, 32, 32)  & 897   &      InputLayer   \\             
                                                                                                  
 Conv2D2      &       (None, 32, 32, 32) &  9249   &     Conv2D1 \\

 Conv2D3      &       (None, 32, 32, 32) &  9249   &     Conv2D2 \\
 Add1             &       (None, 32, 32, 32) &  0    &       Conv2D1 ,  Conv2D3 \\                                                  
                                                                                                  
 Conv2D4         &    (None, 32, 32, 32) &  9249  &      Add1  \\
 Conv2D5         &    (None, 32, 32, 32) &  9249  &      Conv2D4  \\
                                                                                                  
 Add2             &       (None, 32, 32, 32) &  0    &       Add1 ,  Conv2D5 \\                   
 Conv2D6         &    (None, 32, 32, 32) &  9249  &      Add1  \\
 Conv2D7         &    (None, 32, 32, 32) &  9249  &      Conv2D4  \\
                                                                                                  
 Add3             &       (None, 32, 32, 32) &  0    &       Add2 ,  Conv2D7 \\

 Flatten         &   (None, 32768)  &      0    &      Add3 \\
                                                                                                  
 Dense1        &      (None, 128)   &       4194433 &    Flatten \\              
                                                                                                  
 Dense2         &      (None, 10)   &        1290   &     Dense1\\
 \botrule
\end{tabular}
\footnotetext{
Total params: 4,252,187\\
Trainable params: 4,252,187\\
Non-trainable params: 0
}
\end{minipage}
\end{center}
\end{table}

\section{Results}\label{sec4}
In this section, the activation functions are used in two different types of neural networks namely Simple CNN and Residual CNN, and their performances are compared.
 \subsection{Results on Simple CNN}
The CNN model is trained using the above-mentioned hyperparameters on two different datasets namely MNIST \cite{mnist} and CIFAR-10 \cite{cifar} and the results are reported. Two different learning rates are used with the CIFAR-10 dataset and it is shown in Table~\ref{tab2}, and for the MNIST dataset in Table~\ref{tab22}. The proposed TaLU activation function achieves comparatively better performance in this tests with other activation functions.

\begin{table}[!htbp]
\begin{center}
\begin{minipage}{\textwidth}
\caption{Comparison among different activation functions with TaLU for Simple CNN.  Dataset: CIFAR-10\cite{cifar}}\label{tab2}
\begin{tabular*}{\textwidth}{@{\extracolsep{\fill}}lcccc@{\extracolsep{\fill}}}
\toprule%
& \multicolumn{2}{@{}c@{}}{0.01\footnotemark[1]} & \multicolumn{2}{@{}c@{}}{0.1} \\\cmidrule{2-3}\cmidrule{4-5}%
Method & Training & Testing & Training & Testing \\
\midrule
TaLU & 85.14 & \textbf{69.61} & Exploding & 10.00 \\
ReLU & 75.99 & 66.86 & 9.89 & 10.00 \\
SELU & 83.99 & 69.29 & Exploding & 10.00 \\
GELU & 78.29 & 64.84 & Exploding & 10.00 \\
ELU & 83.13 & 69.13 & Exploding & 10.00 \\
Swish & 77.00 & 63.35 & 99.76 & 74.02 \\
LeakyReLU & 77.55 & 68.49 & Exploding & 10.00 \\
SoftPlus & 68.89 & 60.60 & Exploding & 10.00 \\
SoftSign & 64.44 & 63.22 & 100.00 & \textbf{76.61} \\
\midrule
TaLU+BN & 100.00 & \textbf{72.62} & 100.00 & 80.80 \\
ReLU+BN & 100.00 & 70.38 & 100.00 & 80.00 \\
SELU+BN & 100.00 & 72.11 & 100.00 & \textbf{81.13} \\
GELU+BN & 100.00 & 69.65 & 100.00 & 79.89 \\
ELU+BN & 100.00 & 71.96 & 100.00 & 80.06 \\
Swish+BN & 100.00 & 70.80 & 100.00 & 80.24 \\
LeakyReLU+BN & 100.00 & 72.02 & 100.00 & 80.67 \\
SoftPlus+BN & 100.00 & 72.56 & 100.00 & 79.69 \\
SoftSign+BN & 100.00 & 67.53 & 97.88 & 74.28 \\

\botrule
\end{tabular*}
\footnotetext[1]{For the case of without batch normalization, many of the activation function enabled neural network was exploding. So, we set 0.001 learning rate and to converge properly, we increase the epoch from 25 to 50.}
\end{minipage}
\end{center}
\end{table}

\begin{table}[!htbp]
\begin{center}
\begin{minipage}{\textwidth}
\caption{Comparison among different activation functions with TaLU for Simple CNN. Dataset: MNIST\cite{mnist}}\label{tab22}
\begin{tabular*}{\textwidth}{@{\extracolsep{\fill}}lcccc@{\extracolsep{\fill}}}
\toprule%
& \multicolumn{2}{@{}c@{}}{0.01} & \multicolumn{2}{@{}c@{}}{0.1} \\\cmidrule{2-3}\cmidrule{4-5}%
Method & Training & Testing & Training & Testing \\
\midrule
TaLU & 100.00 & \textbf{99.31} & Exploding & 9.80 \\
ReLU & 100.00 & 99.16 & Exploding & 9.80 \\
SELU & Exploding & 9.80 & Exploding & 9.80 \\
GELU & 100.00 & 99.05 & Exploding & 9.80 \\
ELU & 100.00 & 99.23 & Exploding & 9.80 \\
Swish & 100.00 & 99.18 & Exploding & 9.80 \\
LeakyReLU & 100.00 & 99.08 & Exploding & 9.80 \\
SoftPlus & 99.86 & 98.95 & Exploding & 9.80 \\
SoftSign & 99.49 & 99.17 & 100.00 & \textbf{99.32} \\
\midrule
TaLU+BN & 100.00 & \textbf{99.23} & 100.00 & 99.39 \\
ReLU+BN & 100.00 & 99.10 & 100.00 & 99.45 \\
SELU+BN & 100.00 & 99.06 & 100.00 & 99.42 \\
GELU+BN & 100.00 & 99.01 & 100.00 & 99.29 \\
ELU+BN & 100.00 & 99.09 & 100.00 & \textbf{99.47} \\
Swish+BN & 100.00 & 99.12 & 100.00 & 99.45 \\
LeakyReLU+BN & 100.00 & 99.08 & 100.00 & 99.33 \\
SoftPlus+BN & 100.00 & 99.21 & 100.00 & 99.42 \\
SoftSign+BN & 100.00 & 98.92 & 100.00 & 99.23 \\
\botrule
\end{tabular*}
\end{minipage}
\end{center}
\end{table}

\subsection{Results on Residual type Neural Network}
The investigation is also performed on a residual-type neural network. The architecture of the network is shown in Table~\ref{architecture2}. We used the same hyperparameters as simple CNN and the same datasets for the experiments. For simple CNN, comparable or better performance has been for the proposed activation function on both the datasets. The experimental results are reported in Table~\ref{res2} for CIFAR10 dataset, and in Table ~\ref{res22} for MNIST dataset.

\begin{table}[!htbp]
\begin{center}
\begin{minipage}{\textwidth}
\caption{Comparison among different activation functions with TaLU for Residual CNN.  Dataset: CIFAR-10\cite{cifar}}\label{res2}
\begin{tabular*}{\textwidth}{@{\extracolsep{\fill}}lcccc@{\extracolsep{\fill}}}
\toprule%
& \multicolumn{2}{@{}c@{}}{0.01} & \multicolumn{2}{@{}c@{}}{0.1} \\\cmidrule{2-3}\cmidrule{4-5}%
Method & Training & Testing & Training & Testing \\
\midrule
TaLU & Exploding & 10.00 & Exploding & 10.00 \\
ReLU & 91.692 & 60.06 & Exploding & 10.00 \\
SELU & Exploding & 10.00 & Exploding & 10.00 \\
GELU & 100.00 & 62.16 & Exploding & 10.00 \\
ELU & Exploding & 10.00 & Exploding & 10.00 \\
Swish & 100.00 & 60.88 & Exploding & 10.00 \\
LeakyReLU & Exploding & 10.00 & Exploding & 10.00 \\
SoftPlus & 46.904 & 44.89 & Exploding & 10.00 \\
SoftSign & 97.394 & \textbf{64.35} & 10.002 & 9.99 \\
\midrule
TaLU+BN & 100.00 & 67.86 & Exploding & 10.00 \\
ReLU+BN & 100.00 & 68.67 & Exploding & 10.00 \\
SELU+BN & 100.00 & 64.80 & Exploding & 10.00 \\
GELU+BN & 100.00 & 68.47 & Exploding & 10.00 \\
ELU+BN & 100.00 & 66.66 & Exploding & 10.00 \\
Swish+BN & 100.00 & 67.31 & Exploding & 10.00 \\
LeakyReLU+BN & 100.00 & \textbf{69.89} & Exploding & 10.00 \\
SoftPlus+BN & 99.994 & 62.80 & Exploding & 10.00 \\
SoftSign+BN & 100.00 & 65.69 & 99.97 & \textbf{63.16} \\
\botrule
\end{tabular*}
\end{minipage}
\end{center}
\end{table}

\begin{table}[!htbp]
\begin{center}
\begin{minipage}{\textwidth}
\caption{Comparison among different activation functions with TaLU for Residual CNN. Dataset: MNIST\cite{mnist}}\label{res22}
\begin{tabular*}{\textwidth}{@{\extracolsep{\fill}}lcccc@{\extracolsep{\fill}}}
\toprule%
& \multicolumn{2}{@{}c@{}}{0.01} & \multicolumn{2}{@{}c@{}}{0.1} \\\cmidrule{2-3}\cmidrule{4-5}%
Method & Training & Testing & Training & Testing \\
\midrule
TaLU & 100.00 & 98.95 & Exploding & 9.80 \\
ReLU & 100.00 & 99.05 & 100.00 & 97.76 \\
SELU & Exploding & 9.80 & Exploding & 9.80 \\
GELU & 100.00 & 99.07 & Exploding & 9.80 \\
ELU & 100.00 & \textbf{99.14} & Exploding & 9.80 \\
Swish & 100.00 & 98.88 & Exploding & 9.80 \\
LeakyReLU & 99.988 & 98.96 & Exploding & 9.80 \\
SoftPlus & 11.237 & 11.35 & Exploding & 9.80 \\
SoftSign & 99.725 & 98.78 & 99.987 & \textbf{98.98} \\
\midrule
TaLU+BN & 100.00 & \textbf{99.15} & 11.237 & 11.35 \\
ReLU+BN & 100.00 & 99.11 & Exploding & 9.80 \\
SELU+BN & 99.993 & 99.07 & Exploding & 9.80 \\
GELU+BN & 100.00 & 99.04 & 11.237 & 11.35 \\
ELU+BN & 100.00 & 99.09 & Exploding & 9.80 \\
Swish+BN & 100.00 & 99.11 & Exploding & 9.80 \\
LeakyReLU+BN & 100.00 & 99.11 & Exploding & 9.80 \\
SoftPlus+BN & 99.997 & 99.07 & Exploding & 9.80 \\
SoftSign+BN & 99.980 & 99.00 & 99.997 & \textbf{99.17} \\
\botrule
\end{tabular*}
\end{minipage}
\end{center}
\end{table}

\subsection{Result Analysis}
From the tables in the result section, it can be seen that TaLU performs similar or better than other activation functions in terms of achieving accuracy and reducing error rate. It is amongst the top contender to be picked if we want to maximize performance and minimize error rate in the deep neural network training. However, without Batch Normalization, or with a very high learning rate, TaLU can't perform to its potential. With a combination of Batch Normalization and smaller learning rate, TaLU performs impressively and tops the chart with accuracy in most of the cases. Therefore a recommendation to always use Batch Normalization is made as it speeds up learning \cite{batch} and helps stabilize the network and a smaller learning rate with TaLU to unlock the highest potential of TaLU activation function. 

The activation and loss curve of different activation functions are plotted in \ref{accuracy} and in \ref{loss}. These shows that TaLU is very similar with other activation function and quickly converges. TaLU also leads the accuracy and loss curves after 40 epoch.

\section{Conclusion}\label{sec5}
A hybrid activation function called TaLU has been proposed in this paper which is a combination of ReLU (for positive region of the activation function) and Tanh (for the negative region of the activation function). This activation function performs very well in terms of accuracy as it can handle the complex struucture dataset. For result preparation,  when used with Batch Normalization and a smaller learning rate, it can keep up with other activation functions which shows its tremendous potential. This activation function is applied in two different real world datasets and compared the results with other activation functions. Based on the resultant analysis, it can be said that a recommendation of the use of the proposed Talu activation function. However, this researcher  recommends to use this activation function in combination of Batch Normalization and a smaller learning rate.


\bibliography{sn-bibliography}


\begin{thebibliography}{30}
\ifx \bisbn   \undefined \def \bisbn  #1{ISBN #1}\fi
\ifx \binits  \undefined \def \binits#1{#1}\fi
\ifx \bauthor  \undefined \def \bauthor#1{#1}\fi
\ifx \batitle  \undefined \def \batitle#1{#1}\fi
\ifx \bjtitle  \undefined \def \bjtitle#1{#1}\fi
\ifx \bvolume  \undefined \def \bvolume#1{\textbf{#1}}\fi
\ifx \byear  \undefined \def \byear#1{#1}\fi
\ifx \bissue  \undefined \def \bissue#1{#1}\fi
\ifx \bfpage  \undefined \def \bfpage#1{#1}\fi
\ifx \blpage  \undefined \def \blpage #1{#1}\fi
\ifx \burl  \undefined \def \burl#1{\textsf{#1}}\fi
\ifx \doiurl  \undefined \def \doiurl#1{\url{https://doi.org/#1}}\fi
\ifx \betal  \undefined \def \betal{\textit{et al.}}\fi
\ifx \binstitute  \undefined \def \binstitute#1{#1}\fi
\ifx \binstitutionaled  \undefined \def \binstitutionaled#1{#1}\fi
\ifx \bctitle  \undefined \def \bctitle#1{#1}\fi
\ifx \beditor  \undefined \def \beditor#1{#1}\fi
\ifx \bpublisher  \undefined \def \bpublisher#1{#1}\fi
\ifx \bbtitle  \undefined \def \bbtitle#1{#1}\fi
\ifx \bedition  \undefined \def \bedition#1{#1}\fi
\ifx \bseriesno  \undefined \def \bseriesno#1{#1}\fi
\ifx \blocation  \undefined \def \blocation#1{#1}\fi
\ifx \bsertitle  \undefined \def \bsertitle#1{#1}\fi
\ifx \bsnm \undefined \def \bsnm#1{#1}\fi
\ifx \bsuffix \undefined \def \bsuffix#1{#1}\fi
\ifx \bparticle \undefined \def \bparticle#1{#1}\fi
\ifx \barticle \undefined \def \barticle#1{#1}\fi
\bibcommenthead
\ifx \bconfdate \undefined \def \bconfdate #1{#1}\fi
\ifx \botherref \undefined \def \botherref #1{#1}\fi
\ifx \url \undefined \def \url#1{\textsf{#1}}\fi
\ifx \bchapter \undefined \def \bchapter#1{#1}\fi
\ifx \bbook \undefined \def \bbook#1{#1}\fi
\ifx \bcomment \undefined \def \bcomment#1{#1}\fi
\ifx \oauthor \undefined \def \oauthor#1{#1}\fi
\ifx \citeauthoryear \undefined \def \citeauthoryear#1{#1}\fi
\ifx \endbibitem  \undefined \def \endbibitem {}\fi
\ifx \bconflocation  \undefined \def \bconflocation#1{#1}\fi
\ifx \arxivurl  \undefined \def \arxivurl#1{\textsf{#1}}\fi
\csname PreBibitemsHook\endcsname

\bibitem{bib1}
\begin{bbook}
\bauthor{\bsnm{Minsky}, \binits{M.}},
\bauthor{\bsnm{Papert}, \binits{S.A.}}:
\bbtitle{Perceptrons, Reissue of the 1988 Expanded Edition with a New Foreword
  by L{\'e}on Bottou: An Introduction to Computational Geometry}.
\bpublisher{MIT press}, \blocation{???}
(\byear{2017})
\end{bbook}
\endbibitem

\bibitem{sri1}
\begin{barticle}
\bauthor{\bsnm{Ramzan}, \binits{F.}},
\bauthor{\bsnm{Khan}, \binits{M.U.G.}},
\bauthor{\bsnm{Rehmat}, \binits{A.}},
\bauthor{\bsnm{Iqbal}, \binits{S.}},
\bauthor{\bsnm{Saba}, \binits{T.}},
\bauthor{\bsnm{Rehman}, \binits{A.}},
\bauthor{\bsnm{Mehmood}, \binits{Z.}}:
\batitle{A deep learning approach for automated diagnosis and multi-class
  classification of alzheimer’s disease stages using resting-state fmri and
  residual neural networks}.
\bjtitle{Journal of medical systems}
\bvolume{44},
\bfpage{1}--\blpage{16}
(\byear{2020})
\end{barticle}
\endbibitem

\bibitem{sri2}
\begin{barticle}
\bauthor{\bsnm{Najjartabar~Bisheh}, \binits{M.}},
\bauthor{\bsnm{Nasiri}, \binits{G.R.}},
\bauthor{\bsnm{Esmaeili}, \binits{E.}},
\bauthor{\bsnm{Davoudpour}, \binits{H.}},
\bauthor{\bsnm{Chang}, \binits{S.I.}}:
\batitle{A new supply chain distribution network design for two classes of
  customers using transfer recurrent neural network}.
\bjtitle{International Journal of System Assurance Engineering and Management}
\bvolume{13}(\bissue{5}),
\bfpage{2604}--\blpage{2618}
(\byear{2022})
\end{barticle}
\endbibitem

\bibitem{relu}
\begin{bchapter}
\bauthor{\bsnm{Nair}, \binits{V.}},
\bauthor{\bsnm{Hinton}, \binits{G.E.}}:
\bctitle{Rectified linear units improve restricted boltzmann machines}.
In: \bbtitle{Proceedings of the 27th International Conference on Machine
  Learning (ICML-10)},
pp. \bfpage{807}--\blpage{814}
(\byear{2010})
\end{bchapter}
\endbibitem

\bibitem{sri3}
\begin{barticle}
\bauthor{\bsnm{Saha}, \binits{S.}},
\bauthor{\bsnm{Nagaraj}, \binits{N.}},
\bauthor{\bsnm{Mathur}, \binits{A.}},
\bauthor{\bsnm{Yedida}, \binits{R.}},
\bauthor{\bsnm{HR}, \binits{S.}}:
\batitle{Evolution of novel activation functions in neural network training for
  astronomy data: habitability classification of exoplanets}.
\bjtitle{The European Physical Journal Special Topics}
\bvolume{229}(\bissue{16}),
\bfpage{2629}--\blpage{2738}
(\byear{2020})
\end{barticle}
\endbibitem

\bibitem{sri4}
\begin{botherref}
\oauthor{\bsnm{Szanda{\l}a}, \binits{T.}}:
Review and comparison of commonly used activation functions for deep neural
  networks.
Bio-inspired neurocomputing,
203--224
(2021)
\end{botherref}
\endbibitem

\bibitem{sri5}
\begin{barticle}
\bauthor{\bsnm{Farzad}, \binits{A.}},
\bauthor{\bsnm{Mashayekhi}, \binits{H.}},
\bauthor{\bsnm{Hassanpour}, \binits{H.}}:
\batitle{A comparative performance analysis of different activation functions
  in lstm networks for classification}.
\bjtitle{Neural Computing and Applications}
\bvolume{31},
\bfpage{2507}--\blpage{2521}
(\byear{2019})
\end{barticle}
\endbibitem

\bibitem{sri6}
\begin{bchapter}
\bauthor{\bsnm{Graves}, \binits{A.}},
\bauthor{\bsnm{Eck}, \binits{D.}},
\bauthor{\bsnm{Beringer}, \binits{N.}},
\bauthor{\bsnm{Schmidhuber}, \binits{J.}}:
\bctitle{Biologically plausible speech recognition with lstm neural nets}.
In: \bbtitle{Biologically Inspired Approaches to Advanced Information
  Technology: First International Workshop, BioADIT 2004, Lausanne,
  Switzerland, January 29-30, 2004, Revised Selected Papers 1},
pp. \bfpage{127}--\blpage{136}
(\byear{2004}).
\bcomment{Springer}
\end{bchapter}
\endbibitem

\bibitem{relu_usage}
\begin{botherref}
\oauthor{\bsnm{Agarap}, \binits{A.F.}}:
Deep learning using rectified linear units (relu).
arXiv preprint arXiv:1803.08375
(2018)
\end{botherref}
\endbibitem

\bibitem{relu_usage2}
\begin{bchapter}
\bauthor{\bsnm{Hara}, \binits{K.}},
\bauthor{\bsnm{Saito}, \binits{D.}},
\bauthor{\bsnm{Shouno}, \binits{H.}}:
\bctitle{Analysis of function of rectified linear unit used in deep learning}.
In: \bbtitle{2015 International Joint Conference on Neural Networks (IJCNN)},
pp. \bfpage{1}--\blpage{8}
(\byear{2015}).
\doiurl{10.1109/IJCNN.2015.7280578}
\end{bchapter}
\endbibitem

\bibitem{relu_usage3}
\begin{barticle}
\bauthor{\bsnm{{Bai, Yuhan}}}:
\batitle{Relu-function and derived function review}.
\bjtitle{SHS Web Conf.}
\bvolume{144},
\bfpage{02006}
(\byear{2022}).
\doiurl{10.1051/shsconf/202214402006}
\end{barticle}
\endbibitem

\bibitem{sri7}
\begin{barticle}
\bauthor{\bsnm{Zou}, \binits{D.}},
\bauthor{\bsnm{Cao}, \binits{Y.}},
\bauthor{\bsnm{Zhou}, \binits{D.}},
\bauthor{\bsnm{Gu}, \binits{Q.}}:
\batitle{Gradient descent optimizes over-parameterized deep relu networks}.
\bjtitle{Machine learning}
\bvolume{109},
\bfpage{467}--\blpage{492}
(\byear{2020})
\end{barticle}
\endbibitem

\bibitem{sri8}
\begin{bchapter}
\bauthor{\bsnm{Bak}, \binits{S.}},
\bauthor{\bsnm{Tran}, \binits{H.-D.}},
\bauthor{\bsnm{Hobbs}, \binits{K.}},
\bauthor{\bsnm{Johnson}, \binits{T.T.}}:
\bctitle{Improved geometric path enumeration for verifying relu neural
  networks}.
In: \bbtitle{Computer Aided Verification: 32nd International Conference, CAV
  2020, Los Angeles, CA, USA, July 21--24, 2020, Proceedings, Part I 32},
pp. \bfpage{66}--\blpage{96}
(\byear{2020}).
\bcomment{Springer}
\end{bchapter}
\endbibitem

\bibitem{frelu}
\begin{botherref}
\oauthor{\bsnm{Qiu}, \binits{S.}},
\oauthor{\bsnm{Xu}, \binits{X.}},
\oauthor{\bsnm{Cai}, \binits{B.}}:
FReLU: Flexible Rectified Linear Units for Improving Convolutional Neural
  Networks
(2018)
\end{botherref}
\endbibitem

\bibitem{leakyrelu}
\begin{bchapter}
\bauthor{\bsnm{Maas}, \binits{A.L.}},
\bauthor{\bsnm{Hannun}, \binits{A.Y.}},
\bauthor{\bsnm{Ng}, \binits{A.Y.}}, \betal:
\bctitle{Rectifier nonlinearities improve neural network acoustic models}.
In: \bbtitle{Proc. Icml},
vol. \bseriesno{30},
p. \bfpage{3}
(\byear{2013}).
\bcomment{Atlanta, Georgia, USA}
\end{bchapter}
\endbibitem

\bibitem{prelu}
\begin{botherref}
\oauthor{\bsnm{He}, \binits{K.}},
\oauthor{\bsnm{Zhang}, \binits{X.}},
\oauthor{\bsnm{Ren}, \binits{S.}},
\oauthor{\bsnm{Sun}, \binits{J.}}:
Delving Deep into Rectifiers: Surpassing Human-Level Performance on ImageNet
  Classification
(2015)
\end{botherref}
\endbibitem

\bibitem{rrelu}
\begin{botherref}
\oauthor{\bsnm{Xu}, \binits{B.}},
\oauthor{\bsnm{Wang}, \binits{N.}},
\oauthor{\bsnm{Chen}, \binits{T.}},
\oauthor{\bsnm{Li}, \binits{M.}}:
Empirical Evaluation of Rectified Activations in Convolutional Network
(2015)
\end{botherref}
\endbibitem

\bibitem{elu}
\begin{botherref}
\oauthor{\bsnm{Clevert}, \binits{D.-A.}},
\oauthor{\bsnm{Unterthiner}, \binits{T.}},
\oauthor{\bsnm{Hochreiter}, \binits{S.}}:
Fast and Accurate Deep Network Learning by Exponential Linear Units (ELUs)
(2016)
\end{botherref}
\endbibitem

\bibitem{selu}
\begin{botherref}
\oauthor{\bsnm{Klambauer}, \binits{G.}},
\oauthor{\bsnm{Unterthiner}, \binits{T.}},
\oauthor{\bsnm{Mayr}, \binits{A.}},
\oauthor{\bsnm{Hochreiter}, \binits{S.}}:
Self-Normalizing Neural Networks
(2017)
\end{botherref}
\endbibitem

\bibitem{swish}
\begin{botherref}
\oauthor{\bsnm{Elfwing}, \binits{S.}},
\oauthor{\bsnm{Uchibe}, \binits{E.}},
\oauthor{\bsnm{Doya}, \binits{K.}}:
Sigmoid-Weighted Linear Units for Neural Network Function Approximation in
  Reinforcement Learning
(2017)
\end{botherref}
\endbibitem

\bibitem{sri9}
\begin{bchapter}
\bauthor{\bsnm{Laurent}, \binits{C.}},
\bauthor{\bsnm{Pereyra}, \binits{G.}},
\bauthor{\bsnm{Brakel}, \binits{P.}},
\bauthor{\bsnm{Zhang}, \binits{Y.}},
\bauthor{\bsnm{Bengio}, \binits{Y.}}:
\bctitle{Batch normalized recurrent neural networks}.
In: \bbtitle{2016 IEEE International Conference on Acoustics, Speech and Signal
  Processing (ICASSP)},
pp. \bfpage{2657}--\blpage{2661}
(\byear{2016}).
\bcomment{IEEE}
\end{bchapter}
\endbibitem

\bibitem{sri10}
\begin{barticle}
\bauthor{\bsnm{Rajeev}, \binits{R.}},
\bauthor{\bsnm{Samath}, \binits{J.A.}},
\bauthor{\bsnm{Karthikeyan}, \binits{N.}}:
\batitle{An intelligent recurrent neural network with long short-term memory
  (lstm) based batch normalization for medical image denoising}.
\bjtitle{Journal of medical systems}
\bvolume{43},
\bfpage{1}--\blpage{10}
(\byear{2019})
\end{barticle}
\endbibitem

\bibitem{batch}
\begin{botherref}
\oauthor{\bsnm{Ioffe}, \binits{S.}},
\oauthor{\bsnm{Szegedy}, \binits{C.}}:
Batch Normalization: Accelerating Deep Network Training by Reducing Internal
  Covariate Shift
(2015)
\end{botherref}
\endbibitem

\bibitem{gelu}
\begin{botherref}
\oauthor{\bsnm{Hendrycks}, \binits{D.}},
\oauthor{\bsnm{Gimpel}, \binits{K.}}:
Gaussian Error Linear Units (GELUs)
(2020)
\end{botherref}
\endbibitem

\bibitem{softplus}
\begin{bchapter}
\bauthor{\bsnm{Zheng}, \binits{H.}},
\bauthor{\bsnm{Yang}, \binits{Z.}},
\bauthor{\bsnm{Liu}, \binits{W.}},
\bauthor{\bsnm{Liang}, \binits{J.}},
\bauthor{\bsnm{Li}, \binits{Y.}}:
\bctitle{Improving deep neural networks using softplus units}.
In: \bbtitle{2015 International Joint Conference on Neural Networks (IJCNN)},
pp. \bfpage{1}--\blpage{4}
(\byear{2015}).
\doiurl{10.1109/IJCNN.2015.7280459}
\end{bchapter}
\endbibitem

\bibitem{sigmoid}
\begin{barticle}
\bauthor{\bsnm{Narayan}, \binits{S.}}:
\batitle{The generalized sigmoid activation function: Competitive supervised
  learning}.
\bjtitle{Information Sciences}
\bvolume{99}(\bissue{1}),
\bfpage{69}--\blpage{82}
(\byear{1997}).
\doiurl{10.1016/S0020-0255(96)00200-9}
\end{barticle}
\endbibitem

\bibitem{softsign}
\begin{botherref}
\oauthor{\bsnm{Serengil}, \binits{S.I.}}:
Softsign as a Neural Networks Activation Function.
[Online; accessed 2023-05-04]
(2017).
\url{https://sefiks.com/2017/11/10/softsign-as-a-neural-networks-activation-function}
\end{botherref}
\endbibitem

\bibitem{cifar}
\begin{botherref}
\oauthor{\bsnm{Krizhevsky}, \binits{A.}},
\oauthor{\bsnm{Hinton}, \binits{G.}}, et al.:
Learning multiple layers of features from tiny images
(2009)
\end{botherref}
\endbibitem

\bibitem{mnist}
\begin{barticle}
\bauthor{\bsnm{Deng}, \binits{L.}}:
\batitle{The mnist database of handwritten digit images for machine learning
  research}.
\bjtitle{IEEE Signal Processing Magazine}
\bvolume{29}(\bissue{6}),
\bfpage{141}--\blpage{142}
(\byear{2012})
\end{barticle}
\endbibitem

\bibitem{sgd}
\begin{botherref}
\oauthor{\bsnm{Ruder}, \binits{S.}}:
An overview of gradient descent optimization algorithms.
arXiv preprint arXiv:1609.04747
(2016)
\end{botherref}
\endbibitem

\end{thebibliography}


\end{document}